\documentclass{ar-1col}
\pdfoutput=1
\usepackage{linguex}
\usepackage{natbib}
\usepackage{url}
\usepackage{color,colortbl}
\usepackage{array}
\usepackage{multicol}
\usepackage{caption}
\usepackage{subcaption}
\usepackage{xspace}

\definecolor{lightgray}{gray}{0.8}

\newcommand{\ra}{$\rightarrow$ }

\newcommand{\ds}{distributional semantics\xspace}
\newcommand{\Ds}{Distributional semantics\xspace}

\makeatletter%
\def\filter@val(#1)\@nil{\gdef\@currentlabel{#1}}
\def\exi#1{\item[#1]\expandafter\filter@val#1\@nil}
\makeatother

\jname{Annu. Rev. Linguist.}
\jvol{6}
\jyear{2020}
\doi{10.1146/annurev-linguistics-011619-030303}

\begin{document}

\markboth{Boleda}{Distributional Semantics and Linguistic Theory}

\title{Distributional Semantics and Linguistic Theory}

\author{Gemma Boleda$^{1,2}$
  \affil{$^1$Department of Translation and Language Sciences, Universitat Pompeu Fabra, Barcelona 08018, Spain; email: gemma.boleda@upf.edu\\
$^2$Catalan Institution for Research and Advanced Studies (ICREA), Barcelona 08010, Spain\\\\
\normalsize{\textit{\textbf{When citing this paper, please use the following}}:\\ Boleda, G. 2020. Distributional Semantics and Linguistic Theory. \textit{Annu. Rev. Linguist.} 6:213--34. \\
DOI: 10.1146/annurev-linguistics-011619-030303}\\\\
Posted with permission from the Annual Review of Linguistics, Volume 6 \copyright\xspace 2020 by Annual
Reviews, \url{http://www.annualreviews.org/}.}}

\begin{abstract}
Distributional semantics provides multi-dimensional, graded, empirically induced word representations that successfully capture many aspects of meaning in natural languages, as shown in a large body of work in computational linguistics; yet, its impact in theoretical linguistics has so far been limited. This review provides a critical discussion of the literature on distributional semantics, with an emphasis on methods and results that are of relevance for theoretical linguistics, in three areas: semantic change, polysemy and composition, and the grammar-semantics interface (specifically, the interface of semantics with syntax and with derivational morphology).
The review aims at fostering greater cross-fertilization of theoretical and computational approaches to language, as a means to advance our collective knowledge of how it works.


\end{abstract}

\begin{keywords}
distributional semantics, vector space models, vector spaces, semantic spaces, 
computational semantics, semantic change, diachronic semantics,
polysemy, composition, syntax-semantics interface, derivational morphology
\end{keywords}
\maketitle

\tableofcontents

\section{INTRODUCTION}
\label{sec:intro}
This survey provides a critical discussion of the literature on distributional semantics, with an emphasis on methods and results that are of relevance for theoretical linguistics, in three areas: semantic change, polysemy and composition, and the grammar-semantics interface.

Distributional semantics has proven useful in computational linguistics and cognitive
science~\cite[and subsequent work]{landauer-dumais97,schutze92dimensions}; yet, its impact in theoretical linguistics has so far been limited.
A greater cross-fertilization of theoretical and computational approaches promises to advance our knowledge of how language works, and fostering such cross-fertilization is the ultimate goal of this survey.
Accordingly, I will cover mostly research within computational linguistics, rather than cognitive science.


\subsection{Distributional semantics in a nutshell}
\label{sec:ds}

Here I provide only a brief introduction to \ds, such that the survey is self-contained;
for more comprehensive introductions, see \citet{Erk2012}, \citet{clark2015vector}, and \citet{lenci2018distributional}.
\Ds is based on the Distributional Hypothesis, which states that similarity in meaning results in similarity of linguistic distribution~\citep{harris54}: Words that are semantically related, such as \textit{post-doc} and \textit{student}, are used in similar contexts (\textit{a poor \_}, \textit{the \_ struggled  through  the  deadline}; examples from \citeauthor{boleda-herbelot16} \citeyear{boleda-herbelot16}, p.\ 623).
Distributional semantics reverse-engineers the process, and induces semantic representations from contexts of use. 

\begin{figure}[htb]
\centering
\begin{tabular}{ m{0.5\textwidth} m{0.5\textwidth}}
  \centering
  \begin{tabular}{p{0.5\textwidth}}
    Any grad student or \textbf{post-doc} he'd have would be a clonal copy of himself.\\
    During that \textbf{post-doc}, I didn't publish much.\\
    \dots\\
    \centering
    \Large{$\Downarrow$}\\
    \vspace{0.2cm}
    \centering
    \normalsize
      \begin{tabular}{|lcc|}
        \hline
        & dim1 & dim2\\
        \hline
        {\bf post-doc}     &0.71038 & 1.76058\\
        {\bf student}      &0.43679 & 1.93841\\
        {\bf wealth}      &1.77337 & 0.00012\\
        \hline
      \end{tabular}
  \end{tabular}

  & 
\includegraphics[width=2.2in]{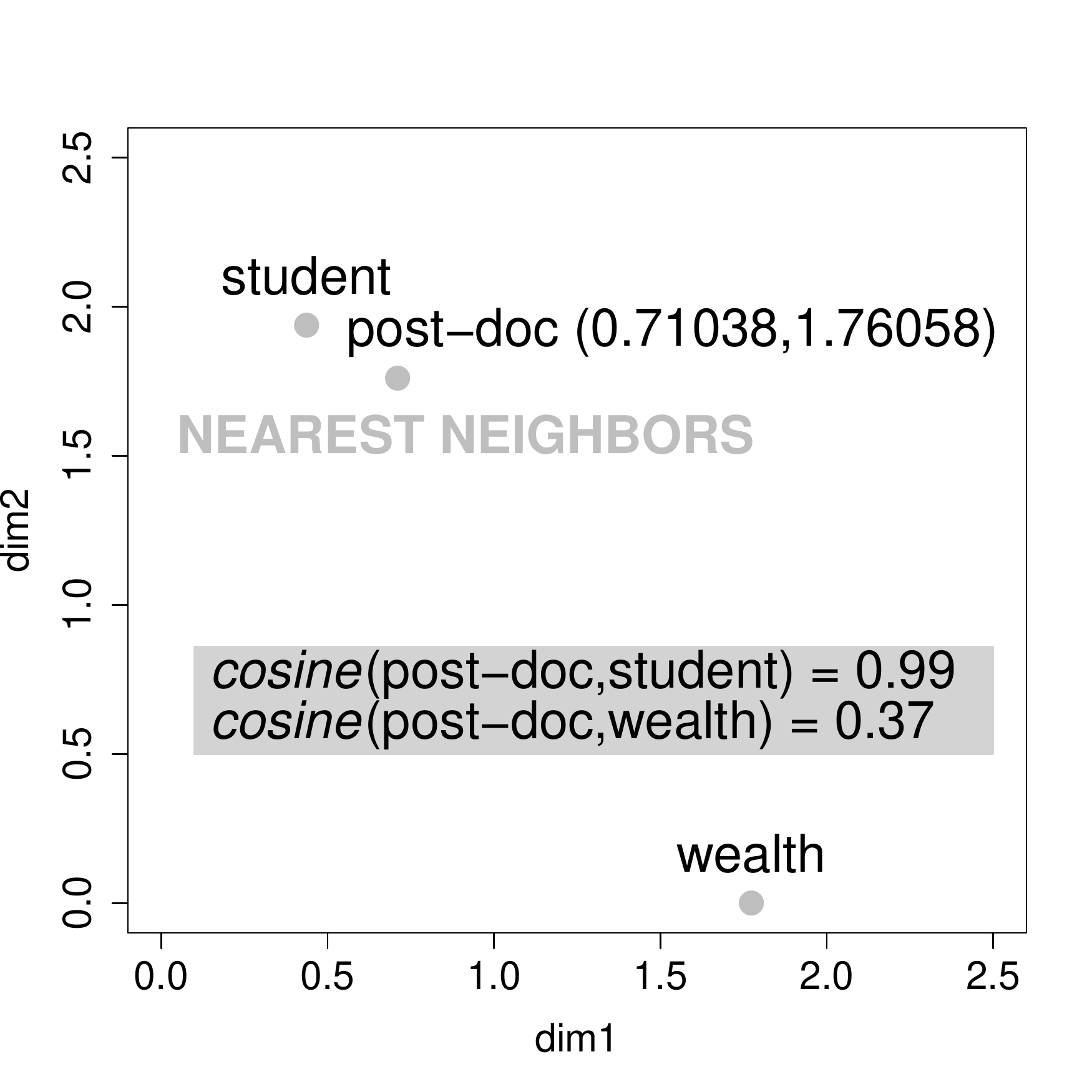}\\
\end{tabular}

\caption{Distributional semantics in a nutshell: Inducing semantic representations from natural language data (left); visualizing and operating with these representations (right). Words are points in a space determined by the values in the dimensions of their vectors, like 0.71028 and 1.76058 for \textit{post-doc}.
  \textit{Post-doc} and \textit{student} are nearer in semantic space than \textit{post-doc} and \textit{wealth}, and in fact they are nearest neighbors of (words closest to) each other.
Adapted from \citet[Figure 1; CC-BY]{boleda-herbelot16}.}
\label{fig:intro-ds}
\end{figure}

In its most basic and frequent form, illustrated in Figure~\ref{fig:intro-ds}, \ds represents word meaning by taking large amounts of text as input and, through an abstraction mechanism (symbolized by the arrow), producing a distributional model, akin to a lexicon, with semantic representations in the form of vectors ---essentially, lists of numbers that determine points in a multi-dimensional space (see below).
However, many more possibilites are available and have been experimented with: The definition of \ds encompasses all kinds of contexts, including for instance the visual context in which words are used~\citep{Baroni2016:grounding}; some models take morphemes, phrases, sentences, or documents instead of words as units to represent~\citep{turney-pantel2010}; and units can be represented via more complex algebraic objects than vectors, such as matrices or tensors~\citep{Grefenstette2011experimenting}.

The collection of units in a distributional model constitutes a vector space or semantic space, in which semantic relations can be modeled as geometric relations.
Vectors determine points in space, and the graph in Figure~\ref{fig:intro-ds} (right) is a graphical rendering of our toy lexicon.
The vectors for \textit{post-doc} and \textit{student} are closer in the space than those of \textit{post-doc} and \textit{wealth}, because their vector values are more similar.
The abstraction mechanisms used to obtain distributional models are such that similar contexts of use result in similar vectors; therefore, vector similarity correlates with distributional similarity, which in turn correlates with semantic similarity, or more generally semantic relatedness.
The most common similarity measure in \ds is the cosine of the angle between two vectors:
the closer the vectors, the larger the cosine similarity.
For instance, the cosine between \textit{post-doc} and \textit{student} in our space is 0.99, while it is 0.37 for \textit{post-doc} vs.\ \textit{wealth} (cosine values for positive vectors range between 0 and 1).

Our example is two-dimensional, but in actual distributional models many more dimensions are used, 300-400 being a frequent range.
While we cannot represent so many dimensions visually, the geometric properties of two-dimensional spaces that we discuss here apply to any number of dimensions.
Given that real distributional vectors are not directly interpretable, a very common way for researchers to gain insight into the information encoded in word vectors is to inspect their nearest neighbors.
These are the words that are closest to a given target; for instance, \textit{student} is the nearest neighbor of \textit{post-doc} in our mini semantic space.

Finally, there are many different versions of the abstraction function (the arrow in Figure~\ref{fig:intro-ds}).
Earlier distributional models were built by extracting and transforming co-occurrence statistics, while recently models based on neural networks have gained ground due to their good performance~\citep{baroni+14:dont}.
Neural networks are a versatile machine learning type of algorithm, used for tasks like machine translation or image labeling; for reasons of scope, in this survey we cover only uses of neural networks that are specifically targeted at building semantic spaces akin to those in classic \ds.

\subsection{Distributional semantics as a model of word meaning}
\label{sec:ds-how}

Distributional semantics largely arises from structuralist traditions~\citep{sahlgren2008distributional}. 
As in structuralism, words are defined according to their position in a system, the lexicon, based on a set of features; their values are defined by contrasts in the words' contexts of use.
However, in structuralism usually only a few features are used, they are defined manually, and they have an intrinsic meaning; for instance, they can be semantic primitives of the sort \textsc{$\pm$male}.
As \cite{boleda2015distributional} point out, in \ds the individual features lack an intrinsic meaning and what gains prominence are the geometric relationships between the words.
Semantic notions like \textsc{$\pm$male} are instead captured in a distributed fashion, as varying patterns across the whole vector.
There are three further key differences to traditional feature-based approaches in linguistics that render \ds attractive as a model of word meaning.

First, the fact that distributional representations are \underline{learnt} from natural language data, and thus radically empirical.
The induction process is automatic, scaling up to very large vocabularies and any language or domain with enough linguistic data to process; for instance, \citet{Bojanowski2017} provide semantic spaces for 157 languages, built from Wikipedia text.
This provides semantic representations on a large scale, in a single, coherent system where systematic explorations are possible. 

\begin{table}[h]
  \tabcolsep7.5pt
  \caption{Near-synonyms in semantic space: The words closest to
    \textit{man}, \textit{chap}, \textit{lad}, and
    \textit{guy} in the distributional model of \protect
    \citet{baroni+14:dont}, adapted from
    \protect \citet{Baroni2016composes}.}
  \label{tab:near-synonyms}
  \begin{center}
    \begin{tabular}{@{}l|l@{}}
      \hline
      Word & Nearest neighbors\\
      \hline
      man & woman, gentleman, gray-haired, boy, person\\
      lad & boy, bloke, scouser, lass, youngster\\
      chap & bloke, guy, lad, fella, man\\
      guy & bloke, chap, doofus, dude, fella\\
      \hline
    \end{tabular}
  \end{center}
\end{table}

Second, high \underline{multi-dimensionality}. The information abstracted from the data is distributed across all the dimensions of a vector, typically a few hundred, which allows for rich and nuanced information to be encoded.
In traditional approaches, again for methodological and practical reasons, comparatively few features are specified. Semantic distinctions can be very subtle, as shown by the phenomenon of near-synonymy. All the words in Table~\ref{tab:near-synonyms} (\textit{man/lad/chap/guy}) denote male adult humans, but each presents different nuances that are difficult to express in a symbolic system using few features.
Their nearest neighbors illustrate the capacity of distributional semantic models to capture both generic and specific semantic features:
On the one hand, most of the neighbors are human- or male-denoting words, suggesting that information akin to semantic features in decompositional approaches, like \textsc{$\pm$male}, is captured in the space (\citeauthor{mikolov-yih-zweig:2013:NAACL-HLT} \citeyear{mikolov-yih-zweig:2013:NAACL-HLT} provide quantitative evidence); on the other hand, the nearest neighbors reflect semantic differences between them, like \textit{lad} being used for younger men (its closest word in the space is \textit{boy}, and it is also near \textit{lass}, used to refer to girls in some English dialects, and \textit{youngster}).

Third, and relatedly, \underline{gradedness}. The information in the vectors is expressed in the form of continuous values, and measures such as cosine similarity are graded: Two vectors can be more or less similar, or similar in certain dimensions but not others.
In the example in Table~\ref{tab:near-synonyms}, even if all four words are near-synonyms, \textit{chap} and \textit{guy} are ``nearer near-synonyms'', if we go by the standard test for synonymy in linguistics \citep[substitutability in context;][]{lyons1977semantics}.
Correspondingly, their vectors are the closest of the set, as shown by their sharing many nearest neighbors.

\section{SEMANTIC CHANGE}
\label{sec:sem-change}
Diachronic semantics, especially lexical semantic change, is an area where the interaction between use and meaning (crucial for \ds) has traditionally been the focus of interest already in theoretical linguistics~\citep{traugott-dasher01,deo2015diachronic}.
For instance, the word \textit{gay} gradually changed during the 20th century from a meaning similar to `cheerful' to its current predominant use as `homosexual', and
its contexts of use in language reflect this change: Examples in~\ref{ex:gay1900} are from the year 1900, and those in~\ref{ex:gay2000} from 2000 (source: COHA, \citeauthor{davies-COHA} \citeyear{davies-COHA}).
The three key properties of \ds mentioned above are useful to model semantic change, and this is currently a blooming topic in computational linguistics~\cite[for overviews, see][]{KutuzovEtal-coling2018,Tahmasebi2018}: High dimensionality allows it to represent many semantic nuances that can be subject to change, gradedness in representations is crucial to account for the gradual nature of change, and, as we will see below, its data-driven nature allows it to detect semantic change from changes in usage.

\ex.\label{ex:gay1900}
She was a fine-looking woman, cheerful and gay.\\
We assembled around the breakfast with spirits as gay and appetites as sharp as ever.

\ex.\label{ex:gay2000}
[\dots] the expectation that effeminate men and masculine women are more likely to be seen as gay men and lesbians, respectively. \\
`I don't personally support gay marriage myself,' Edwards said.

\subsection{Distributional approaches to diachronic semantics}

Distributional methods started being used for semantic change around the 2010s, with initial works using classic distributional methods \citep{sagi2009,gulordava2011distributional} and \citet{kim2014temporal} introducing neural network representations, which have been predominant in later work~\citep{hamilton2016diachronic,szymanski2017temporal,deltredici+19stms}.
Distributional approaches are based on the hypothesis that a change in context of use mirrors a change in meaning, which can be seen as a special case of the Distributional Hypothesis.
They thus infer a change in meaning when they observe a change in the context of use.

This is typically done by building word representations at different points in time and comparing them (although \citeauthor{rosenfeld-erk-2018-deep} \citeyear{rosenfeld-erk-2018-deep} include time as a variable in the model instead).
This method is used to both detect semantic change and track its temporal evolution.
For instance, \citet{kim2014temporal} built one distributional lexicon per year from 1850 to 2009 using data from the Google Book Ngrams corpus~\citep{michel2011googlecorpus}.
The cosine similarity of the word \textit{gay}, when compared to its representation in the 1900 lexicon, goes down through the 20th century, with the drop accelerating at the end of the 70s from around 0.75 to around 0.3 in 2000.

\begin{figure}[h]
\includegraphics[width=\textwidth]{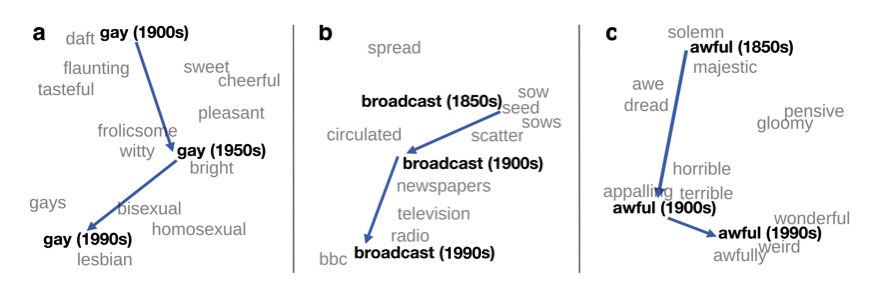}
\caption{Two-dimensional visualization of semantic change for three English words, reproduced from \citet{hamilton2016diachronic}. The figure was obtained via dimensionality reduction from a space with 300-dimensional vectors. Reproduced with permission.}
\label{fig:gay}
\end{figure}

Figure~\ref{fig:gay} visualizes the trajectory of three words across time in another study \citep{hamilton2016diachronic}, with nearest neighbors in gray font along the words of interest.
It illustrates how inspection of nearest neighbors can help trace the specific meaning shift taking place.
In 1900, \textit{gay} is near words like \textit{daft} or \textit{cheerful}, and by 1990 it is instead near to \textit{homosexual}. 
the change in \textit{broadcast} is metaphorical in nature, from a concrete to a more abstract meaning (from spreading seeds to spreading information or signal); and \textit{awful} undergoes pejoration, from a positive to a negative denotation.
Another method used to track specific semantic changes is targeted comparisons to words related to the old and the new meanings: For instance, \citet{kim2014temporal} compare how the cosine similarities of \textit{cell} to \textit{dungeon} and \textit{phone} evolve through the years.
However, the latter requires previous knowledge of the specific change taking place.

There is current experimentation on two related efforts \citep{Tahmasebi2018}: sense-specific semantic change, where sense representations are induced and then tracked (also see Section~\ref{sec:wsenses}), and detecting not only the presence but also the type of semantic shift.
In the latter literature, starting with the pioneering work of \cite{sagi2009}, there is some evidence that distributional methods can spot narrowing and broadening, two classically described types of diachronic shift~\citep{hock91}.
A case of narrowing is `deer', which evolved from Old English \textit{deor}, meaning `animal', to its current narrower denotation; one of broadening is \textit{dog}, from Late Old English \textit{docga}, which used to denote a specific breed of dog, to its current broader meaning.
An extreme form of broadening results in grammaticalization, as in verb \textit{do} going from a lexical to an auxiliary verb between the 15th and the 18th century.
\cite{sagi2009} trace these three words by representing each context of use individually, with one vector per sentence.
They show that, for \textit{dog} and \textit{do}, contexts become more separate over time, corresponding to the broadening effect, and the reverse for \textit{deer}.
Moreover, their distributional measure 
correlates with the proportion of periphrastic uses of \textit{do} through the centuries, independently estimated via manual annotation of texts.

Up to now, most research has focused on showing \underline{that} \ds can model semantic change, rather than on systematically exploring data and advancing our knowledge of the phenomenon.
An exception is~\citet{xu2015computational}, a study assessing two previously proposed laws that make contradicting predictions.
Their large-scale computational analysis, based on distributional semantic models of English on the Google Ngrams corpus, shows that pairs of synonyms tend to stay closer in space than control pairs across the 20th century, in four dataset jointly comprising tens of thousands of words.
They thus provide support for the law of parallel change~\citep{stern21}, that posits that related words undergo similar changes, and against the law of differentiation~\citep{breal97}, that defends that synonyms tend to evolve different meanings because it is not efficient for languages to maintain synonyms.
Other generalizations about semantic change emerging from work with distributional methods have been proposed, but controlled experiments have called them into question~\citep{dubossarsky-etal-2017-outta}.

\subsection{Discussion}

\Ds has tremendous potential to accelerate research in semantic change, in particular the exploration of large-scale diachronic data, in four main crucial points: (1)~detecting semantic change, as a change in the representation of a word across time; (2)~temporally locating it, by monitoring the rate of change in the distributional representation; (3)~tracking the specific semantic evolution of the word, via an inspection of the nearest neighbors or targeted examination of cosine similarities; (4)~testing competing theories via large-scale empirical studies.
It can also help detecting the type of semantic change, although this is still an under-researched topic.

A major challenge is the fact that distributional methods, especially those based on neural networks, are quite data-hungry, while many datasets in diachronic semantics are rather small~\citep{KutuzovEtal-coling2018}.
This means that most studies are for English, and other languages are neglected: Of 23 datasets used for diachronic semantics, identified in \citet{Tahmasebi2018}'s survey, only 4 are not in English.
Moreover, the vast majority of studies focus on the Google Book Ngrams corpus, which covers only the 1850-2009 period.

When the amount of data is scarce, spurious effects easily arise.
For instance, \citet{deltredici+19stms}, in a study of meaning shift in a community of soccer fans with data from 2011 to 2017, find that reference to specific people or events causes changes in cosine similarity that do not correspond to semantic change; an example is \textit{stubborn}, which in 2017 was mostly used when talking about a new coach.%
\footnote{\citet{deltredici+19stms} detect such cases with a complementary distributional measure, based on the specificity of the contexts of use.}
Effects like this challenge the Distributional Hypothesis, as a change in context does not signal a change in meaning, and call for more nuanced methods.
This kind of issue is typically less problematic for studies involving longer time scales, because of the larger amount and variety of data, but it can arise when data are scarce or when there are systematic differences in the sources for different time periods ---for instance if texts are from different genres.

Another issue is that research has mostly focused on lexical semantic change, while in diachronic semantics there is much work on grammaticalization processes~\citep{deo2015diachronic}.
While classic distributional approaches could not account for function words (to the point that they were typically removed from the vocabulary), recent neural network models do provide usable representations for them~\citep{Mikolov2010,peters-etal-2018-deep}, opening new possibilities.



\section{POLYSEMY AND COMPOSITION}
\label{sec:pol}

Words are notoriously ambiguous or polysemous, that is, they adopt different meanings in different contexts~\citep[among many others]{cruse86}.
For instance, \textit{post-doc} refers to a person in the first sentence in Figure~\ref{fig:intro-ds}, and to a period of time in the second.
\Ds has traditionally tackled this issue in two ways, which resonate with linguistic treatments of polysemy~\citep{lyons1977semantics}. The predominant approach, by far, is to take the word as a unit of representation and provide a single representation that encompasses all its uses (Section~\ref{sec:comp}).
The second approach is to provide different vectors for different word senses (Section~\ref{sec:wsenses}).

\subsection{Single representation, polysemy via composition}
\label{sec:comp}

The predominant, single representation approach is similar in spirit to structured approaches to the lexicon like the Generative Lexicon \citep{pustejovsky95}, Frame Semantics \citep{fillmore06frame}, or HPSG~\citep{pollard1994head}, even if not directly inspired by them. 
These approaches aim at encoding all the relevant information in the lexical entry, and then define mechanisms to deploy the right meaning in context, usually by composition.
As an example, \citet[122-123]{pustejovsky95} formalizes two readings of \textit{bake}, a change of state (\textit{John baked the potato}) and a creation sense (\textit{John baked the cake}), by letting the lexical entries of the verb and the noun interact: If \textit{bake} combines with a mass-denoting noun, the change of state sense emerges; if it combines with an artifact, the creation sense emerges.
This has the advantage of capturing aspects of meaning that are common to the different contexts, while being able to account for the differences.
Sense accounts of polysemy struggle with this, and face a host of other serious theoretical, methodological, and empirical issues~\citep[see][for discussion]{pustejovsky95,kilgarriff97}.

\begin{figure}[h]
  \begin{tabular}{ m{0.4\textwidth} m{0.6\textwidth}}
  \centering
    \textbf{corpus-based}\\
  \begin{tabular}{|lcc|}
    \hline
     & dim1 & dim2\\
    \hline
    {\bf cut}     &4 & 5\\
    {\bf cost}      &1 & 5\\
    \textbf{cut cost}      &4 & 9\\
    \hline
  \end{tabular}

    \vspace{0.2cm}
    \textbf{synthetic}\\
    \begin{tabular}{|lcc|}
    \hline
     & dim1 & dim2\\
    \hline
    \textsc{cut cost}      &5 & 10\\
    \hline
  \end{tabular}

     & \includegraphics[width=2.2in]{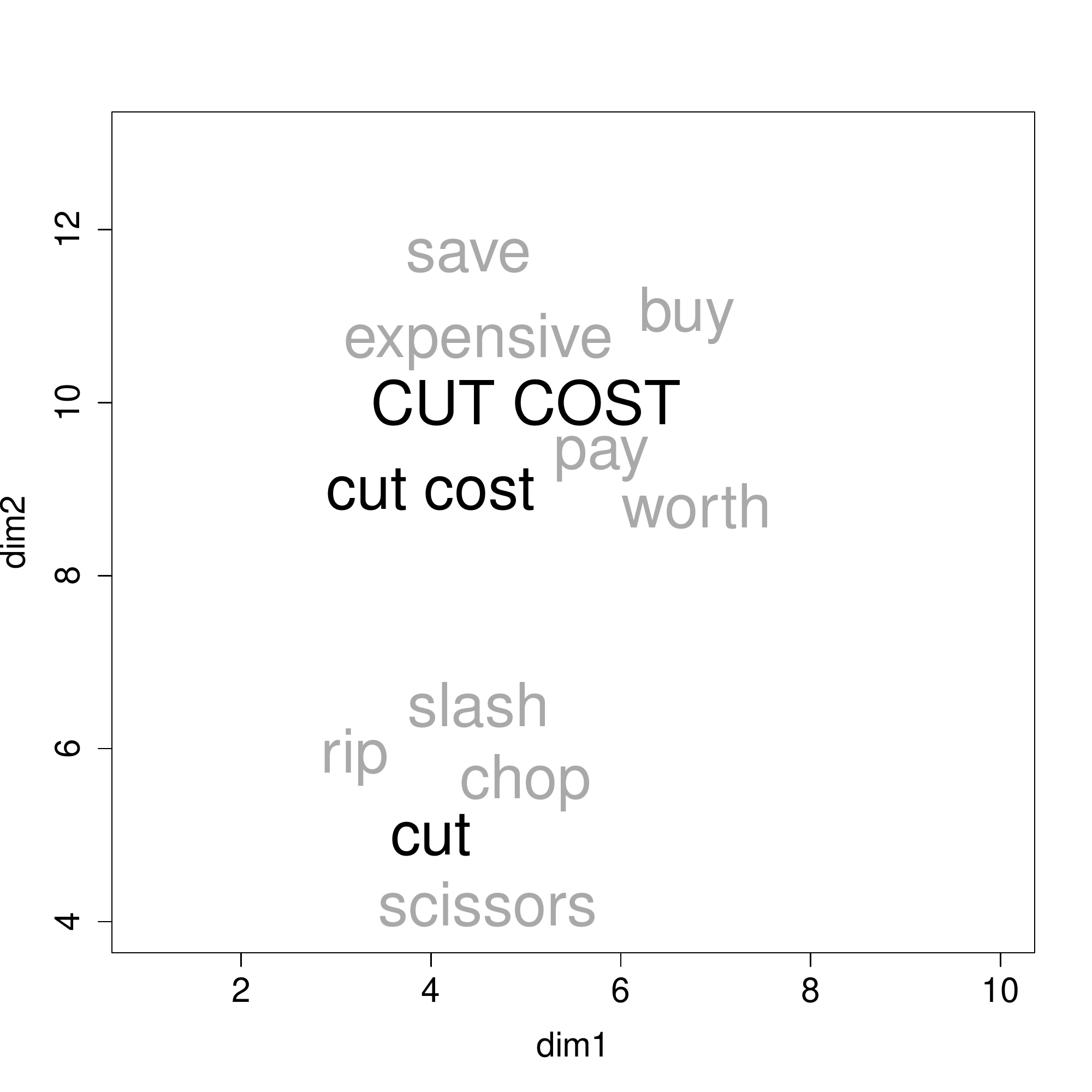}\\
  \end{tabular}
  \caption{Compositional distributional semantics: Illustration with vector addition. Left: The synthetic vector \textsc{cut cost} is built by component-wise addition of the vectors for \textit{cut} and \textit{cost}. Right: The argument \textit{cost} pulls the vector for \textit{cut} towards its abstract use (see nearest neighbors, in gray). The corpus-based vector for \textit{cut cost} can be used to check the quality of its synthetic counterpart.}
  \label{fig:composition}
\end{figure}

In standard \ds, each word is assigned a single vector, which is an abstraction over all its contexts of use, thus encompassing all the word senses that are attested in the data~\citep{arora+2018-linear}.
The pioneering work of~\citet{kintsch2001} in Cognitive Science started extending distributional methods to compose such generic word representations into larger constituents.
The computational linguistic community took this research line up almost a decade later \citep{erk-pado08structured,mitchell-lapata2010,baroni2010nouns,coecke2011mathematical,socher12:comp,mikolov2013distributed}.
Compositional distributional methods build representations for phrases out of the representations of their parts, and the corresponding meaning is expected to emerge as a result of the composition~\citep{baroni+13:Frege}.
Figure~\ref{fig:composition} provides an illustration, with the simplest composition method: adding the word vectors.
The synthetic vector \textsc{cut cost} created via this composition method has a value of 4+1=5 for the first dimension (\textit{dim1}) because the values of \textit{cut} and \textit{cost} for this dimension are 4 and 1, respectively.

To give an intuition of how this may account for semantic effects, let's assume that dimension 1 is associated to abstract notions and dimension 2 to concrete notions (of course this is a simplification; remember that properties like concreteness are captured in a distributed fashion).
The verb \textit{cut} has a concrete sense, as in \textit{cut paper}, and a more abstract sense akin to \textit{save}, as in \textit{cut costs}, and so it has high values for both dimensions. Instead, \textit{cost} is an abstract notion, and so it has low values for dimension 1 and high values for dimension 2.
When composing the two, the abstract dimension gets highlighted, pulling the vector towards regions in the semantic space related to its abstract sense.
This is shown in Figure~\ref{fig:composition} (right); while the vector values are fictitious, the neighbors (in gray) are a selection of the 20 nearest neighbors of \textit{cut} and \textsc{cut cost} in a real semantic space~\citep{mandera2017explaining}.%
\footnote{User-friendly interface to this semantic space: \url{http://meshugga.ugent.be/snaut-english}.}
As the nearest neighbors show, the representation of \textit{cut} is dominated by the physical sense, but its composition with \textit{cost} shifts it towards the abstract sense.
This mechanism by which matching semantic dimensions reinforce each other, and mismatched dimensions remain less active, is reminiscent of the mechanisms by \citet{pustejovsky95} discussed above for \textit{bake a potato} vs.\ \textit{bake a cake}.
The main difference is that distributional representations are not explicitly structured like those in the Generative Lexicon, although that does not mean that they lack structure; rather, the structure is implicitly defined in the space.

A substantial body of work has shown that composition methods in \ds largely account for polysemy effects in semantic composition.
\citet{baroni2010nouns} and subsequent work compare the synthetic vector for a phrase, like \textsc{cut cost} in Figure~\ref{fig:composition}, to a phrase vector \textit{cut cost} that is extracted directly from the corpus with standard distributional methods.
The closer the synthetic vectors are to the corpus-based ones, the better the composition method.
In~\citet{boleda+13intensionality}, the best composition method obtains an average cosine similarity of 0.6 between synthetic and corpus-based vectors for adjective-noun phrases; for comparison, phrases have an average cosine similarity of 0.4 to their head nouns.
Another common method is to compare model results to human intuitions about the semantics of phrases.
\citet{mitchell-lapata2010} introduced this for phrase similarity (in turn inspired on methods to evaluate word similarity), with participant data such as \textit{reduce amount - cut cost} being very similar, \textit{encourage child - leave company} very dissimilar, and \textit{present problem - face difficulty} obtaining medium scores.
The best composition methods yield Spearman correlation scores with participant data around 0.4 (minimum is 0, maximum 1) for adjective-noun, noun-noun, and verb-noun phrases; for comparison, correlation scores between different participants are around 0.5.
Other work experiments with ditransitive constructions~\citep{Grefenstette2011experimenting}, with triples such as \textit{medication achieve result - drug produce effect}, or even full sentences \citep{bentivogli2016sick}, but going beyond short phrases proves difficult.
There is not much work on function words because, as mentioned above, these are traditionally hard to model with \ds.
An exception is \citet{bernardi-EtAl:2013:Short}, which seeks to identify paraphrasing relationships between determiner phrases (e.g., \textit{several wives}) and words that lexically involve some form of quantification (e.g., \textit{polygamy}).
They obtain reasonable but not optimal results.


A particularly exciting application of compositional distributional methods is that of~\citet{Vecchi2017}, who showed that distributional models are able to distinguish between semantically acceptable and unacceptable adjective-noun phrases.
Crucially, their data involves phrases that are unattested in a very large corpus; some phrases are unattested because they are semantically anomalous (\textit{angry lamp}, \textit{legislative onion}), and some due to the generative capacity of language, with its explosion of combinatory properties, together with the properties of the world, which make some combinations of adjectives and nouns unlikely even if they are perfectly acceptable (\textit{warm garlic}, \textit{sophisticated senator}).
The fact that distributional models are able to predict which combinations are acceptable for human participants, and which are not, suggests that they are able to truly generalize.

Work in this area has investigated much more sophisticated approaches to composition than vector addition.
I cannot do justice to this research for reasons of space, but see \citet{Erk2012} and \citet{baroni2013composition}.
Much of this work is inspired by formal semantics~\citep{garrette2011integrating,Beltagy:13montague,baroni+13:Frege,coecke2011mathematical,lewis2013combined,Herbelot2015building,Erk2016alligator};
\citet{boleda-herbelot16} surveys research at the intersection between formal and \ds.
However, a robust result that has emerged from all this literature is that vector addition is surprisingly good, often outperforming more sophisticated methods.
This suggests that the greatest power of \ds lies in the lexical representations themselves.

Relatedly, this research has also shown that, while \ds can model composition of content words in short phrases, scaling up to larger constituents and accounting for function words remains challenging.
Recall that \ds provides abstractions over all occurrences of an expression.
Compositionally built phrases remain generic rather than grounded to a specific context.
Therefore, distributional approaches can account for the fact that, in general, \textit{red box} will be used for boxes that are red in color, but they cannot really account for highly context-dependent interpretations, like \textit{red box} referring to a brown box containing red objects~\citep{mcnally-boleda17}.
This is because \ds does not come equipped with a mechanism to integrate word meaning in a given linguistic and extralinguistic context, or to represent that context in the first place~\citep{aina+19}.
Note that functional elements like tense or determiners need a context to be interpreted, so it makes sense that they are challenging for \ds.

Accordingly, \citet{westera-boleda19} defend the view that \ds accounts for expression meaning (more concretely, how an expression is typically used by speakers), but not for speaker meaning (how a speaker uses an expression, in terms of communicative intentions, in a given context, cf.\ the \textit{red box} example above).
Newer generation neural networks are contributing to expanding these limits, as they natively incorporate mechanisms to compose new words with a representation of the context~\citep{Mikolov2010}.
However, the extent to which they can account for speaker meaning, and contextual semantic effects more generally, remains to be seen; the results in \cite{aina+19} suggest that some of these models still overwhelmingly rely on lexical information.

Finally, another area where \ds shows potential is the phenomenon of semantic opacity and semi-opacity, which is the opposite end of compositionality; see Section~\ref{sec:morph} for work on the compositionality of noun compounds.

\subsection{Different representations, polysemy via word senses}
\label{sec:wsenses}

Other work aims at building sense-specific distributional representations, where typically each word sense is assigned a different vector~\cite[for a recent survey, see][]{camacho2018word}.
The key insight here is that, because \ds is based on context of use, and uses of a word in a given sense will be more similar to each other than to uses of the same word in a different sense, we can detect word senses by checking the similarity of the contexts.
The pioneering work of~\cite{schuetze98} did this by 
representing single instances of word use, with one vector for each sentence in which the word occurs.
Then, word senses were automatically identified as coherent regions in that space.
This work started a tradition, within \ds, of research on word sense induction and sense-specific word representations~\citep{mccarthy+04finding,reisinger-mooney10multi}.
Erk and colleagues instead aimed at providing a representation of the specific meaning a word takes in a given context, entirely bypassing word senses~\citep{erk-pado08structured,erk-pado10,Erk2013measuring}.
This work is related to compositional distributional semantics, with the difference that it provides a use-specific word vector representation instead of going directly to the representation of the larger constituent.

Two crucial problems in sense-based approaches to polysemy are 1)~deciding when two senses are different enough to warrant the addition of an item to the vocabulary, 2)~how to represent the information that is common to different senses~\citep{kilgarriff97}.
\Ds does not improve things with respect to the first issue, but it does alleviate the second.
Two sense-specific vectors can be similar in some dimensions (e.g., for \textit{cut}, those related to reducing or splitting) and different in others (like the abstract/concrete axis of \textit{cut}), in a graded fashion.
The same way that it can capture similarities and differences between words, it can capture similarities and differences between word senses.

\subsection{Discussion}

Polysemy is a pervasive phenomenon that is difficult to model in a discrete, symbolic system~\citep{kilgarriff97}.
\Ds provides an attractive framework, complementary to traditional ones~\citep{heylen2015monitoring}.
Multi-dimensionality allows it to capture both the common core to different uses of a word and the differential factors, as some dimensions of meaning can specialize in the former and some in the latter.
Gradedness allows it to capture the degree of the semantic shift in different contexts of use, be it in the composition route or the word sense route.
Moreover, the fact that \ds provides data-induced representations for a large number of words makes it possible to use it to make predictions and test specific hypotheses driven by linguistic theory.

As an example,~\cite{boleda+13intensionality} test the hypothesis, stemming from formal semantics, that modification by a certain class of adjectives is more difficult to model than other classes.
The specific prediction is that synthetic phrases with these adjectives (like \textsc{alleged killer}) will be further away from their corpus-based vectors than synthetic phrases with other adjectives (like \textsc{severe pain}).
Their results are negative, and they instead observe the influence of another factor in the results:
If an adjective denotes a very typical property of a noun, like \textit{severe} for \textit{pain}, then it is easy to model; if it is less typical, like \textit{severe} for \textit{budget}, then it is more difficult.
In many of the difficult cases, such as \textit{likely base}, it is not even clear how the two words compose; out of context, it is not easy to come up with possible interpretations for this phrase.
This led the authors to further explore the context-dependence of modification, resulting in a theoretical proposal \citep{mcnally-boleda17}.
The authors propose that composition exploits two aspects of meaning: on the one hand, the conceptual aspect, with regularities in how words match (\textit{box} denotes a physical object, and \textit{red} is typically used to specify colors of physical objects); on the other hand, the referential aspect, specifically the information about the referent of the phrase (for instance, \textit{red box} can be used in a context that requires distinguishing a brown box containing red objects from another, identical-looking brown box containing blue objects).
\Ds can model conceptual, but not referential effects, for the same reason that it cannot model contextual effects and speaker meaning more generally (see Section~\ref{sec:comp}).
\cite{mcnally-boleda17} took distributional semantic data themselves as an object of empirical inquiry; they asked what made certain phrases difficult for compositional distributional models, and the answer proved theoretically worthy.
This work is thus an example of fruitful collaboration between computational and theoretical approaches to language.


\section{GRAMMAR-SEMANTICS INTERFACE}
\label{sec:interfaces}

There is ample evidence that content-related aspects of
language interact with formal aspects, as is salient for
instance in argument structure and the expression of arguments
in syntax~\citep[see Section~\ref{sec:syntax}]{grimshaw1990argument,levin93}, as well as in
derivational morphology~\citep[see Section~\ref{sec:morph}]{lieber2004morphology}.

\subsection{Syntax-semantics interface}
\label{sec:syntax}

Beth Levin's seminal work on the
syntax-semantics interface was based on the observation
that ``the behavior of a verb, particularly with respect to the
expression of its arguments, is to a large extent determined by its
meaning'' \citep[p.\ 1]{levin93}.
She defines semantic verb classes on the basis of several syntactic properties.
This is a particular case of the Distributional Hypothesis, and thus it is
natural to turn it around and use distributional cues to infer semantic classes ---as Levin herself does in her research in a manual fashion.

Levin's work had a big impact in Computational Linguistics, inspiring work on the automatic acquisition of semantic classes from distributional evidence~\citep{dorr-jones96,merlo-stevenson01,mccarthy00,korhonen+03,lapata-brew04,schulteimwalde06,boleda+12:CL}.
For instance, \citet{merlo-stevenson01} used manually defined linguistic features, with data extracted from corpora, 
to classify English verbs into three
optionally transitive classes: unergative, unaccusative and object-drop. They
achieved around 70\% accuracy.
Other work targets a finer-grained classification, with Levin-style semantic classes, such as \citet{schulteimwalde06} for German.
This early work used distributional evidence, but not \ds strictu sensu. 
\citet{baroni-lenci10}
replicated \citet{merlo-stevenson01}'s experiment using a proper
distributional model, obtaining comparable accuracies.

\begin{figure}[h]
\includegraphics[width=3in]{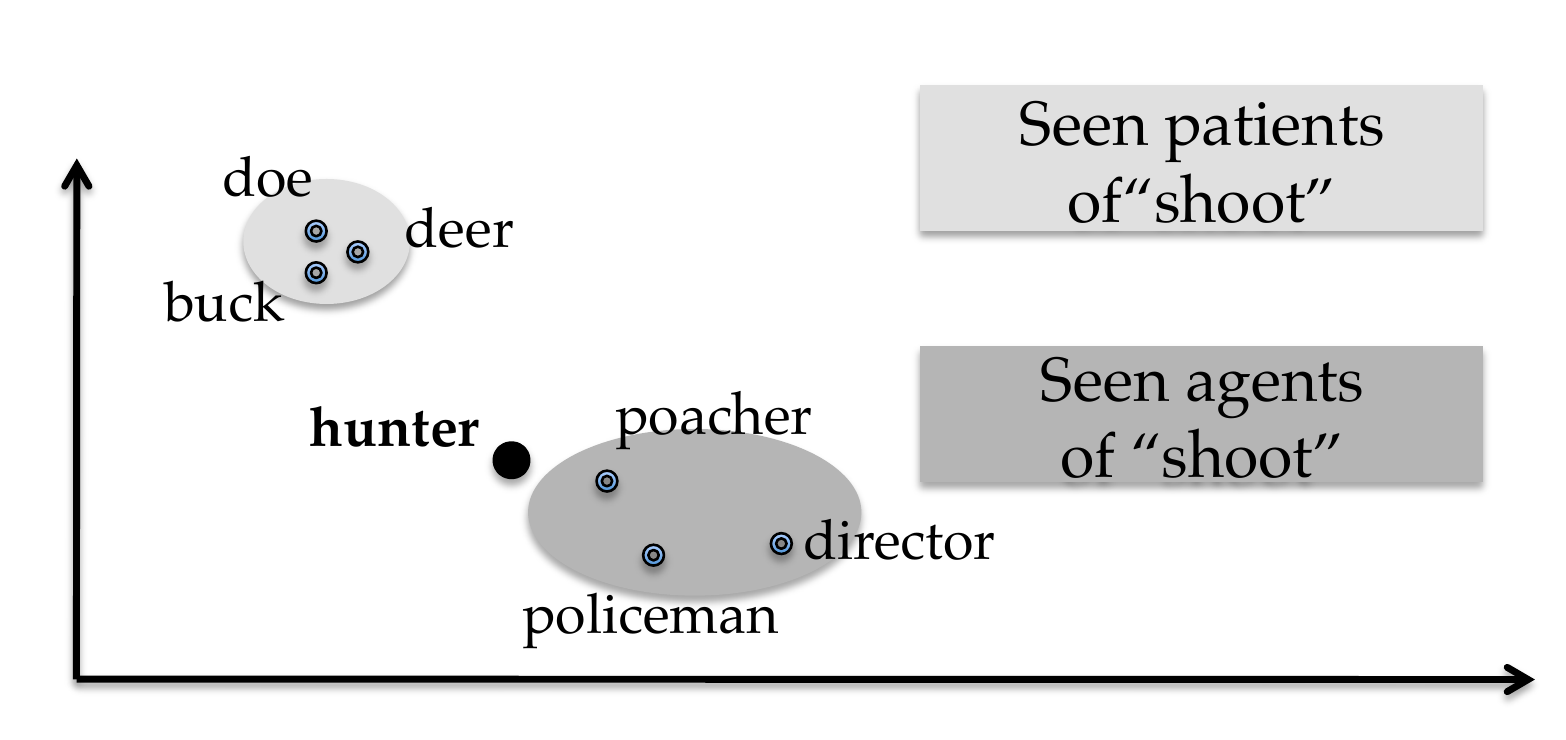}
\caption{Visualization of the approach to selectional preferences in \cite{erk+10}. Partial reproduction of their Figure 1, p.\ 731; CC-BY.}
\label{fig:erk-selpref}
\end{figure}

\citet{erk+10} initiated a line of work using distributional methods to model selectional restrictions, or the thematic fit between an argument and a predicate (usually, a verb).
They capitalize on the fact that distributional models capture gradedness in linguistic phenomena, since selectional restrictions are graded: \emph{cake} is a better object for \emph{eat} than \emph{chalk}, which is in turn better than
\emph{sympathy}.
Again, this is not easy to capture in symbolic models with discrete features like [$\pm$\textsc{edible}].
\citeauthor{erk+10} computed the plausibility of each verb-argument combination as the similarity between a candidate argument and a (weighted) average of the arguments observed with a verb.
For instance, when deciding whether \textit{hunter} is a plausible agent for the verb \textit{shoot}, they computed its similarity to an average of the vectors for \textit{poacher, director, policeman}, etc.\ (see Figure~\ref{fig:erk-selpref}).
This average vector can be seen as a prototype for the argument of the verb.

\citeauthor{erk+10} compared the scores of the model to human ratings (where participants were asked to rate the plausibility that e.g.\ \textit{hunter} is an agent of \textit{shoot}).
Their model achieved a Spearman correlation score of 0.33 and 0.47 ($p<0.001$) with the human ratings in two different datasets for English involving agent and patient roles.
Erk et al.'s idea of working with argument prototypes has been further refined and developed in subsequent models~\citep{baroni-lenci10,Lenci2011,greenberg-etal-2015-improving,santus-etal-2017-measuring} with improved empirical results and a broader coverage of phenomena.





\subsection{Morphology-semantics interface}
\label{sec:morph}

Derivational morphology is at the interface between grammar and
semantics~\citep{lieber2004morphology}.
Stem and affix need to match in both morphosyntactic and semantic features: For instance, the suffix \emph{-er} applies to verbs, as in \textit{carve} \ra \textit{carver}, but only those that have certain kinds of arguments.
The effects of derivational processes are also both 
grammatical (\emph{-er} produces nouns) and semantic (these nouns have some agentive connotation, like \emph{carver}).
Derivational processes are semi-regular; they are largely compositional, but not always (mainly due to lexicalization processes), and they present subregularities (for instance, \emph{carver, driver} denote agents, but \emph{broiler, cutter} denote instruments).
Moreover, both stem and affix semantics exhibit the properties typical of word semantics, such as polysemy and gradedness~\citep{Marelli2015}; cf.\ the polysemy of \textit{-er} between agent and instrument.
Thus, accounting for morphological derivation requires fine-grained lexical semantic representations for both stem and affix and mechanisms to combine them, in a clear analogy to phrase composition (see Section~\ref{sec:comp}).

In recent years, researchers have explored methods to produce distributional representations for morphologically complex words from the representations of their parts~\citep[a.o.]{lazaridou-EtAl:2013:ACL2013,Marelli2015,pado+16coling,lapesa2018disambiguation,cotterell-schutze-2018-joint}; most of this work has adapted compositional methods initially developed for word composition.
The motivation in this work is two-fold. From a theoretical point of view, distributional semantics offers new tools to investigate derivational morphology, in particular its rich, data-driven semantic representations.
From a practical perspective, such methods address ``the data problem'' of distributional semantics \citep[p.\ 1285]{pado+16coling}: In general, morphologically complex words are less frequent than morphologically simple words, and thus distributional representations for morphologically complex words can be expected to be of a comparatively lower quality;
moreover, because morphology is productive, new words are continuously created, and, in these cases, data is simply unavailable, making it imperative to rely on methods to build synthetic word vectors for morphologically complex words.


Researchers have experimented with simple composition methods and more complex ones, often based on machine learning.
Again, the simplest method is addition, which here implies summing up the vectors for the stem and the affix, as in \textsc{carver} = \textit{carve} + \emph{er}.
However, affixes are not observed as units in corpora.
A common method to obtain affix representations is to average derived words~\citep{pado+16coling} ---for instance, averaging the vectors for \textit{carver}, \textit{drinker}, \textit{driver}, etc.\ to obtain a representation for \textit{-er}.

\begin{table}[h]
\tabcolsep7.5pt
\caption{Derivational phenomena captured with compositional distributional semantic methods: Examples from \cite{Marelli2015}.}
\label{tab:morph}
\begin{center}
\begin{tabular}{@{}l|l|l@{}}
\hline
  Phenomenon & Word & Nearest neighbors (selection)$^{\rm a}$\\
\hline
  Affix polysemy & \textsc{carver}$^{\rm b}$ & potter, engraver, goldsmith\\
                 & \textsc{broiler} & oven, stove, to cook, kebab, done\\
\hline
  Sense selection & column & arch, pillar, bracket, numeric\\
                 & \textsc{columnist} & publicist, journalist, correspondent\\
\hline
 Differential effect & \textsc{industrial} & environmental, land-use, agriculture\\
 of the affix &  \textsc{industrious} & frugal, studious, hard-working\\
\hline
\end{tabular}
\end{center}
\begin{tabnote}
  $^{\rm a}$\cite{Marelli2015} provide a selection of the 20 nearest neighbors; 
  $^{\rm b}$in small caps, synthetic word representations, produced by derivation operations with distributional semantics (see text).
\end{tabnote}
\end{table}

Table~\ref{tab:morph} showcases phenomena captured by the distributional model of \citet{Marelli2015}, illustrated through nearest neighbors (see the original paper for quantitative evaluation; also note that their composition method is more sophisticated than addition, but the kinds of effects modeled are similar for different composition functions).
Words in small caps correspond to synthetic word vectors, the rest to corpus-based word vectors.
The first block of the table shows that the distributional method captures the agent/instrument polysemy of the affix, and is able to produce different results depending on the stem:
The synthetic vector for \textsc{carver} is near agents for professions, like \textit{potter} or \textit{goldsmith}, whereas \textsc{broiler} is in the region of cooking instruments (\textit{oven}, \textit{stove}).
In the second block, we see that the relevant sense of the stem is captured even in cases where it is not the predominant one: In the vector for the word \textit{column}, the senses related to architecture and mathematics
  dominate (see nearest neighbors), but \textit{-ist} correctly focuses on the
  sense related to journalism when producing \textsc{columnist}.
  Because \textit{-ist} often produces
  professions, its distributional representation is able to
  select the dimensions of \textit{column} that match one of the meaning
  types produced by the morpheme.
Finally, the examples in the third block show that different affixes produce different meanings when applied to the same stem. For instance, \textsc{-al} and \textsc{-ous} have quite different
  consequences on the same base form.

\Ds has clear potential to capture linguistic phenomena related to derivation; the extent to which they are able to do so is still under investigation, since distributional methods exhibit a large variance in performance across invididual words and across derivational patterns~\citep{pado+16coling}.
  The factors intervening are still not fully understood, but it seems clear that some are methodological and some are linguistic.
  As for the former, if a word is very frequent, it will have probably undergone lexicalization; if it is very unfrequent, then its corpus-based representation will be of low quality.
  In both cases, the word will not be a good candidate to participate in the creation of the affix representation, or as a comparison point to evaluate distributional methods.
  It is thus not surprising that overall scores are good but not optimal.
  For instance, \citet{lazaridou-EtAl:2013:ACL2013}, in a study on English,
  showed that derived forms have a mean cosine similarity of 0.47 with their base forms (e.g.\ \textit{carver} compared to \textit{carve}).
  The best compositional measure provides a mean similarity of 0.56 between synthetic and corpus-based vectors ---significantly higher, but not a big jump.
  However, they also provide evidence that, in cases where the quality of the corpus-based word representations is low, the compositional representation is substantially better, suggesting that distributional methods can provide useful semantic representations for derived words in a productive way, and alleviate the data problem explained above. For instance, the nearest neighbors of \textit{rename} in their distributional space are \textit{defunct, officially, merge}, whereas those for the synthetic vector \textsc{rename} are \textit{name, later, namesake}.

As for linguistic factors, for instance \citet{pado+16coling}, in a large-scale study of derivation in German, find that the derivational pattern is the best predictor for model performance (that is, some derivational processes are intrinsically harder to model than others), and argue that derivations that create new argument structure tend to be harder for distributional models: For instance, the agentive/instrumental nominalization with suffix \textit{-er} (\textit{fahren \ra Fahrer}, English \textit{drive-driver}), where the external argument is incorporated into the word, is difficult to capture, whereas deverbal nominalizations that preserve argument structure are comparatively easy (e.g.\ with suffix \textit{-ung}, \textit{umleiten \ra Umleitung}, English \textit{redirect-redirection}).

Research in derivational morphology also shows that vector addition works surprisingly well, as was the case with composition (see Section \ref{sec:comp}). 
This again suggests that the distributional representations themselves do most of the job, and are more important than the specific method used.
\citet{cotterell-schutze-2018-joint}'s results underscore this interpretation. They propose a probabilistic model that integrates the automatic decomposition of words into morphemes (\textit{carver} \ra \textit{[carve] [er]}) with the synthesis of their word meaning, jointly learning the structural and semantic properties of derivation.
They test different derivation models and different word representations on English and German data, with representations having by far the most influence on the results.

The robustness of addition has emerged also in the study of semantic opacity and semi-opacity, which typically aims at predicting the degree of compositionality in compound nouns and multi-word expressions. 
\citet{reddy2011empirical}, a representative study, aimed at reproducing human ratings on the degree of compositionality of 90 English compound nouns (\textit{climate change}, \textit{graduate student}, \textit{speed limit} obtaining maximum compositionality scores, \textit{silver bullet}, \textit{ivory tower}, \textit{gravy train} minimum).
Adding the two component vectors (with a higher weight of the modifier; see paper for details) achieved a Spearman correlation score of 0.71 with human data. 
Other work uses different methods; for instance, \citet{springorum2013detecting} do not use compositional methods but explore how the modifier and the head contribute to compositionality ratings for German data.
Against their prediction, the modifier is a much better predictor of compositionality than the head.

Again, most work is directed at showing \underline{that} \ds can model derivational morphology, rather than tackling more specific linguistic questions.
An exception is \cite{lapesa2017doggies}, an interdisciplinary collaboration between theoretical and computational linguists that tests hypotheses about the effect of derivation on emotional valence (the positive or negative evaluation of the referent of a word), on German data.
For instance, one of the predictions is that diminutives shift words towards positive valence (consider \textit{Hund} \ra \textit{H\"undchen}, English \textit{dog}-\textit{doggie}).
The work provides overall support for the hypotheses, but in a nuanced form.
The most interesting result is a hitherto unobserved interaction of many valence effects with concreteness:
The diminutive makes nouns positive if they denote concrete objects, whereas it tends to make abstract nouns negative (compare the case of \textit{dog} with \textit{Idee} \ra \textit{Ideechen}, English \textit{idea}-\textit{small idea}), 
and verbal prefixation with \textit{\"uber-} (\textit{over-}) tends to make concrete verbs, but not abstract verbs negative (\textit{fahren} \ra \textit{\"uberfahren}, Eng.\ \textit{drive}-\textit{run over} vs. \textit{nehmen} \ra \textit{\"ubernehmen}, Eng.\ \textit{take}-\textit{take over}).
This work again showcases the potential of \ds to uncover linguistically relevant factors.

Although most work is on derivational morphology, some research has tackled inflection, too. A very influential study, with a model first proposed by \citet{rumelhart1973model}, is~\cite{mikolov-yih-zweig:2013:NAACL-HLT}, which showed that several morphological and semantic relations are organized according to simple additive relationships in distributional space.
For instance, \textit{good}-\textit{better}+\textit{rough} creates a synthetic vector that is very near \textit{rougher}.
The idea is that if you subtract an inflected word from its stem, you obtain a representation of the affix (here, the comparative), which can then be applied to a new stem (here, \textit{rough}) by addition.
\citeauthor{mikolov-yih-zweig:2013:NAACL-HLT} tested eight patterns involving nominal, adjectival, and verbal inflection,
obtaining an average accuracy of 40\% on the task of predicting the missing element in the tuple.
40\% may not look very impressive, but it is if we consider that the model is required to find the exact right answer in a vocabulary of 82,000 words ---that is, in the case above, the answer is counted as correct only if the nearest neighbor of the synthetic vector \textsc{rougher} is the word vector of \textit{rougher}.

\subsection{Discussion}


The work reviewed in this section has three main assets to offer to theoretical approaches to the grammar-semantics interface.
The first is a wealth of data, created as part of the research in order to develop and evaluate distributional methods.
For example, participant ratings on the compositionality of compounds~\citep{reddy2011empirical} can be used when selecting material for experimental research.
Other examples are ratings of typicality and semantic relatedness~\citep{springorum2013detecting,lazaridou-EtAl:2013:ACL2013} or information about derived words, such as derivational pattern and degree of polysemy~\citep{pado+16coling}.
This kind of contribution is common to other quantitative and computational work~\citep{baayen1993celex}.

The second is tools to create and explore data via distributional methods.
For instance, the similarity between a derived form and a combination of its components in distributional space can be used as a proxy for its degree of compositionality, which is useful to explore processes of derivation and lexicalization.
Other linguistic features can be simulated with distributional measures: For instance, \citet{pado+16coling} measure how semantically typical a base form is for a given morphological pattern by comparing it to the average of all the bases in the pattern
(e.g.\, for \textit{-er}, the word vector for \textit{carve} compared to the average of the vectors for \textit{carve, drink, drive, broil, cut}, etc.).

The third is its potential to uncover new empirical facts that are of potential theoretical significance:
For instance, the suggestion in \citet{pado+16coling} that derivation processes that affect argument structure are more challenging to model computationally, or the relevance of the concreteness/abstractness axis in the study of \citet{lapesa2017doggies}.

\section{CONCLUSION AND OUTLOOK}
\label{sec:conc}
The above discussion summarizes robust results in \ds that can be directly imported for research in theoretical linguistics, as well as challenges and open issues.
Among the robust results are that
(1)~\ds is particularly useful in areas where the connection between use, meaning, and grammar is relevant, such as the areas reviewed in this survey;
(2)~geometric relationships in distributional models correspond to semantic relationships in language;
(3)~gradedness in distributional representations correlates with gradedness in semantic phenomena (e.g., the degree of semantic change);
(4)~averaging the distributional representations of classes of words yields useful abstractions of the relevant classes (e.g., of arguments accepted by specific predicates);
(5)~simple combinations of distributional representations produce quite accurate predictions as to the semantics of phrases and derived words.
I have argued that the multi-dimensional, graded, and data-driven nature of its representation are key aspects that contribute to these results.

There are at least four ways for distributional semantic research to contribute to linguistic theories.
The first is exploratory.
Distributional data such as similarity scores and nearest neighbors can be used to explore data on a large scale.
The second is as a tool to identify instances of specific linguistic phenomena. For instance, changes in distributional representations of words across time can be used to systematically harvest potential instances of semantic change in diachronic data (Section~\ref{sec:sem-change}).
The third is as a testbed for linguistic hypotheses, by testing predictions in distributional terms.
The fourth, and hardest, is the actual discovery of linguistic phenomena or theoretically relevant trends in data. This requires collaboration between computational and theoretical linguists.

There are also a number of challenges that distributional methods face.
Like other data-driven methods, distributional models mirror the data they are fed.
This is good, because they provide radically empirical representations, and also dangerous, because representations are subject to biases in the underlying data~\citep{caliskan2017semantics}.
A related challenge is the fact that distributional methods need large amounts of data to learn reasonable representations. A rule of thumb is to have at least 20-50 instances of each expression to represent; many languages, domains, or time periods simply lack these data.
There is active research on faster learning, as this is a problem for many other areas, but no working solution for the moment.
A final, crucial issue is the lack of adequate researcher training, which prevents a wider use of distributional semantics in linguistics, and of quantitative and computational methods more generally.
Strengthening student training in such methods in linguistics degrees is of paramount importance to allow the field to adequately exploit the vast amounts of linguistic data that have become available in the last few decades. 

In this survey, to maximize readability, I have focused on simple methods such as vector similarity, nearest neighbors, vector addition, and vector averaging.
While these are the basic methods in the field, a glaring omission are methods based on machine learning techniques, which are also commonly used to extract information from semantic spaces and operate with distributional representations.
I refer the reader to the references in the survey for more information.

For reasons of scope, I have also left out of the discussion research on neural networks that is not specifically targeted at building semantic spaces.
Neural networks are a type of machine learning algorithm, recently revamped as deep learning~\citep{lecun2015deep}, that induce representations of the data they are fed in the process of learning to perform a task.
For instance, they learn word representations as they learn to translate from English to French, given large amounts of bilingual text. 
They proceed by trial and error, attempting to translate a sentence, measuring the degree of error, and feeding back to the representations such that they become more helpful for the task.
Linguistic tasks that are general enough, like machine translation or word prediction, result in general-purpose representations of language.
Most deep learning systems for language include a module that is akin to a distributional lexicon, and everything I have said in this survey applies to such modules.
However, crucially, these systems also have other modules that represent linguistic context, and mechanisms to combine this context with word representations.
This is a big step with respect to classic distributional models, and deep learning is being adopted in the community at top speed.
To illustrate, the examples in \ref{ex:play1}-\ref{ex:play2} (from \citeauthor{peters-etal-2018-deep} \citeyear{peters-etal-2018-deep}, p. 2233)
show the nearest neighbors of two sentences containing the polysemous word \textit{play}, where 
the representations for the sentences are vectors produced by the complex compositional function implemented in the neural network.
The nearest neighboring sentences illustrate that the model has captured not only the relevant sense of the word, but also more nuanced aspects of the meanings of the sentences (commenting on good plays in \ref{ex:play1}, referring to signing for plays rather than to the acting itself in \ref{ex:play2}).

\ex.\label{ex:play1}
\textit{Sentence}: Chico Ruiz made a spectacular \underline{play} on Alusik's grounder [\dots]\\
\textit{Nearest neighbor}: Kieffer [\dots] was commended for his ability to hit in the clutch, as well as his all-round excellent \underline{play}.

\ex.\label{ex:play2}
\textit{Sentence}: Olivia De Havilland signed to do a Broadway \underline{play} for Garson [\dots]\\
\textit{NN}: [\dots] they were actors who had been handed fat roles in a successful \underline{play} [\dots]

Given the success of these models, and their complexity, there is a booming interest in the computational linguistic community in understanding what aspects of language they capture, and how~\citep{alishahi2019analyzing}.
Recently, \cite{pater2019generative} has argued for the integration of neural network models in linguistic research (also see the responses to his article).
I could not agree more.


\section*{DISCLOSURE STATEMENT}
The author is not aware of any affiliations, memberships, funding, or financial holdings that
might be perceived as affecting the objectivity of this review.

\section*{ACKNOWLEDGMENTS}
I am grateful to Louise McNally, Josep Maria Fontana, Alessandro Lenci, and Marco Baroni for discussions about the role of distributional semantics in linguistic theory, to the AMORE team (Laura Aina, Kristina Gulordava, Carina Silberer, Ionut Sorodoc, and Matthijs Westera) for shaping my current thinking about this topic, and to Dan Jurafsky for a helpful review of a previous manuscript. 
This project has received funding from the European Research Council (ERC) under the European Union's Horizon 2020 research and innovation programme (grant agreement No 715154), and from the Spanish Ram\'on y Cajal programme (grant RYC-2015-18907). 
This paper reflects the author's view only, and the EU is not responsible for any use that may be made of the information it contains.


\bibliographystyle{ar-style1}
\bibliography{gemma-short,not-to-save}

\end{document}